\documentclass[sigconf,10pt]{acmart}
\AtBeginDocument{%
  \providecommand\BibTeX{{%
    \normalfont B\kern-0.5em{\scshape i\kern-0.25em b}\kern-0.8em\TeX}}}

\usepackage{hyperref}
\usepackage{cleveref}
\usepackage{multirow}
\usepackage{array}
\usepackage{makecell}
\usepackage{xcolor}
\usepackage{algorithm}
 \usepackage{algorithmicx}
 \usepackage{algpseudocode}
\usepackage{booktabs} 

\usepackage{amsmath}               
  {
      
  }

\definecolor{darkgreen}{rgb}{0.0, 0.8, 0.0} 
\usepackage{listings}
\algrenewcommand\algorithmicrequire{\textbf{Input:}}
\algrenewcommand\algorithmicensure{\textbf{Output:}}

\usepackage{mathtools}
\DeclarePairedDelimiter{\ceil}{\lceil}{\rceil}
\DeclarePairedDelimiter{\floor}{\lfloor}{\rfloor}

\begin{document}

\title{Tin-Tin: Towards Tiny Learning on Tiny Devices with Integer-based Neural Network Training}

\author{Yi Hu}
\affiliation{%
  \institution{Carnegie Mellon University}
  \country{}
}

\author{Jinhang Zuo}
\affiliation{%
  \institution{City University of Hong Kong}
  \city{}
  \country{}
}
\author{Eddie Zhang}
\affiliation{%
  \institution{Carnegie Mellon University}
  \city{}
  \country{}
}
\author{Bob Iannucci}
\affiliation{%
  \institution{Carnegie Mellon University}
  \city{}
  \country{}
}
\author{Carlee Joe-Wong}
\affiliation{%
  \institution{Carnegie Mellon University}
  \city{}
  \country{}
}





\newcommand{\eve}[1]{{ {\color{blue} #1}}}
\newcommand{\note}[2]{{\color{blue}\tiny{$\downarrow$}\scriptsize{#1\%}} \textbf{#2}}

\newcommand{\carlee}[1]{{\color{red} C: #1}}
\newcommand{\bob}[1]{{\color{teal} B: #1}}
\newcommand{\jinhang}[1]{{\color{orange} J: #1}}

\begin{abstract}
Recent advancements in machine learning (ML) have enabled its deployment on resource-constrained edge devices, fostering innovative applications such as intelligent environmental sensing. However, these devices, particularly microcontrollers (MCUs), face substantial challenges due to limited memory, computing capabilities, and the absence of dedicated floating-point units (FPUs). These constraints hinder the deployment of complex ML models, especially those requiring lifelong learning capabilities. To address these challenges, we propose Tin-Tin, an integer-based on-device training framework designed specifically for low-power MCUs. Tin-Tin introduces novel integer rescaling techniques to efficiently manage dynamic ranges and facilitate efficient weight updates using integer data types. Unlike existing methods optimized for devices with FPUs, GPUs or FPGAs, Tin-Tin addresses the unique demands of tiny MCUs, prioritizing energy efficiency and optimized memory utilization. We validate the effectiveness of Tin-Tin through end-to-end application examples on real-world tiny devices, demonstrating its potential to support energy-efficient and sustainable ML applications on edge platforms.
\end{abstract}




\maketitle
\section{Introduction}
The rapid development of the Internet of Things (IoT) and edge computing has significantly expanded the deployment of machine learning (ML) on edge devices, thereby enhancing the efficiency and responsiveness of applications like wearable sensing~\cite{nyamukuru2020tinyeats,liu2023realtime,jeyakumar2019senseHAR} and smart cities~\cite{khan2020edge,chen2021edge,alahi2023integration}. Moreover, many emerging applications, such as wildfire monitoring~\cite{bushnaq2021role,liu2021sensor} or detecting faults in manufacturing equipment~\cite{rai2021machine,xames2023systematic},
necessitate maintenance-free, long-lived sensor networks~\cite{lorincz2009mercury,hester2017future,sparks2017route} that can deploy ML models to detect anomalies and make predictions about the environment. In these scenarios, low-power sensors and microcontrollers (MCUs) play a critical role, being both cost-effective and disposable, to ensure sustainability and longevity.

However, enabling ML on these resource-constrained devices presents many new challenges. Power-efficient MCUs typically have limited memory and computing capabilities, as detailed in Sec. \ref{sec:memory_mot}. For instance, the popular STM32 Cortex M0\footnote{https://www.st.com/en/microcontrollers-microprocessors/stm32f0-series.html} MCUs have up to 144 KB of data RAM, and the MSP430\footnote{https://www.ti.com/microcontrollers-mcus-processors/msp430-microcontrollers/overview.html} mixed-signal MCUs offer up to 66 KB. In addition to memory constraints, both lack dedicated floating point units (FPUs), which are designed to handle  operations on floating point numbers. Consequently, these MCUs can only perform integer arithmetic or rely on slow software implementations for floating-point calculations. These limitations significantly restrict the complexity of ML models that can be deployed on these edge devices.

Furthermore, most existing research on edge ML focuses only on inference tasks. However, in practical edge applications with evolving environments, on-device lifelong learning~\cite{khouas2024training,zhu2022device} becomes crucial. Devices must dynamically learn new tasks and continuously adapt to evolving input distributions. In manufacturing applications, for example, it is common for real operating conditions to differ from test ones~\cite{liu2023lifelong,chen2022lifelong}, requiring a fault detection model to be re-trained in real time on operational data. Moreover, manufacturing equipment degrades, often in an unpredictable manner, over time; sensors using ML to predict faults, then, must adapt their models to this degradation over time~\cite{ding2024self,zhang2024deep}. Training, unlike inference, imposes significantly higher demands on memory and computational resources (Sec. \ref{sec:memory_mot}, \ref{sec:int-motive}).
For instance, it may require $7\times$ longer compared to inference as demonstrated in \cite{xu2022mandheling}.
This challenge is particularly demanding for tiny edge devices like MCUs, where battery life is a critical concern, necessitating efficient resource management to prevent rapid battery drain during training.

Our goal is to support and optimize \emph{tiny, lifelong} data learning on these \emph{highly constrained} low-power MCUs with only hundreds of KB memory while extending battery life. Efforts like TinyML \cite{unite-ai2023tinyml,banbury2021benchmarking} and model compression techniques \cite{han2015deep,wang2019haq,basu2019qsparse,liberis2021munas} to reduce the energy usage and computational overhead of on-device model deployment 
are effective for inference but cannot handle model updates. Meanwhile, model training approaches using reduced-precision models and weight updates~\cite{bernstein2018signsgd,courbariaux2015binaryconnect,zhu2016trained,li2022ternary} still rely on intensive intermediate floating point computations and may require specialized non-standard hardware.
To the best of our knowledge, \textit{ours is the first integer-based on-device training framework designed specifically for low-power MCUs}. While NITI~\cite{wang2022niti} addresses some basic challenges of integer-based training, it is primarily tailored for GPUs and FPGAs rather than tiny MCUs, leaving several critical questions unanswered for deployments on power-constrained devices, including the two technical challenges that we address below. 
Table \ref{tab:related_compare} offers a comprehensive comparison of our approach to related work. 

To enhance energy efficiency in lifelong learning while satisfying memory and computing constraints, we introduce \textbf{Tin-Tin}, an innovative integer-based on-device training framework for \textbf{Tin}y learning on \textbf{Tin}y MCUs. Specifically, our approach replaces full-precision floating points with reduced-precision low-bitwidth data types, significantly decreasing memory requirements by using smaller data formats (e.g., \texttt{fp32} to \texttt{int8}) and simplifying calculations (e.g., floating point to integer arithmetic). This reduction  complements existing structure- and network-level optimizations like sparse updates~\cite{lin2022ondevice} and pruning~\cite{liu2019metapruning}.
Unlike existing methods optimized for IoT devices with FPUs~\cite{lin2022ondevice} or GPUs and FPGAs~\cite{wang2022niti}, Tin-Tin specifically targets the unique demands of tiny MCUs by prioritizing energy efficiency and optimized memory utilization, effectively addressing two primary \textbf{technical challenges}:

\paragraph{\textit{Restricted Representation Range} }Using integer arithmetic for neural network training requires integer representations of weights ($w$), activations ($a$), gradients ($g$) and errors ($e$). Unlike floating points, which have a wide representation range  (e.g., $1.4$e$^{-45}$ to $3.4$e$^{38}$ for \texttt{fp32}), integers have a limited range (e.g., $[-128,127]$ for \texttt{int8}). Fixed ranges based on presumed value limits can lead to overflow (unwanted clipping) or reduced precision (Sec. \ref{sec:dr_motivation}). To maintain accuracy, value ranges should be dynamically adjusted during training, as the distributions of inputs, outputs, and weights change. This dynamic adjustment could enhance model performance but necessitates an appropriate rescaling scheme throughout the training process, an aspect unaddressed in previous integer-based training methods.

\paragraph{\textit{Integer Weight Update}} Maintaining accurate model updates with integer weights is challenging due to their limited precision. The desired update values often need to be much smaller than the weights themselves. Traditional gradient descent techniques achieve this by using a small learning rate (e.g., 0.001) to scale down the gradient update, guiding the model towards the optimum. However, in integer-based training, directly applying a small learning rate is problematic because multiplying integer gradients by a small scalar produces floating-point values. Converting these scaled floating-point gradients back to integers introduces additional computational costs and precision issues. Therefore, innovative methods are required to apply integer gradient updates that are much smaller in scale than the weights, while maintaining accuracy within the integer-only framework.

To tackle these challenges and target resource-constrained MCUs, we introduce novel integer rescaling techniques that support a dynamic range and efficient weight updates. We demonstrate the effectiveness of our framework through comprehensive end-to-end application examples on tiny devices. Our primary \textbf{contributions} include:
\begin{itemize}
\item We propose Tin-Tin, a pioneering framework tailored for on-device training on MCUs, leveraging integer-based model quantization to significantly reduce memory requirements and improve energy efficiency.
\item We introduce an integer-only rescaling method and a compound exponentiation scaling scheme to provide dynamic range support, ensuring accurate scale management throughout training and minimizing overflow and quantization errors. By dynamically adjusting scaling factors, Tin-Tin maintains training stability and performance on MCUs.
\item We develop an efficient weight update mechanism that aligns gradients and weights to a common scale through integer-based operations, including gradient alignment and adaptive update step sizes, preserving model accuracy and efficiency during training.
\item We evaluate Tin-Tin across various learning applications, focusing on its energy and memory efficiency on different MCUs. Case studies on motor bearing fault detection and spectrum sensing demonstrate Tin-Tin's ability to significantly reduce training time, memory usage, and energy consumption. 
Our real-world experiments on MCUs validates that Tin-Tin can save up to 82\% in energy and 40\% in memory compared to full-precision floating-point implementations.
\end{itemize}

The remainder of the paper is organized as follows: Section~\ref{sec:related} contrasts our work with related studies. Section~\ref{sec:background} provides the background and motivation for our research. Section~\ref{sec:system} introduces the Tin-Tin framework and discusses our novel integer rescaling techniques for managing dynamic ranges and facilitating efficient weight updates.  Section~\ref{sec:eval} presents the experimental validation of our work on various MCUs. Finally, we conclude in Section~\ref{sec:conclusion}.

\begin{table*}
    \caption{Related Work and Comparison.  
    }
    \label{tab:related_compare}
    \begin{center}
    \begin{tabular}{l l l l l l l l }\toprule
         &  \makecell[cl]{Optimization\\Target} & \makecell[cl]{Applied\\Stage} & \makecell[cl]{Pre-runtime\\Data Not\\Required} & \makecell[cl]{Integer\\Backprop} &\makecell[cl]{Dynamic\\Weight\\Range} & \makecell[cl]{Low-Power\\MCU\\Deployment$^{\dagger}$} \\\midrule
       \makecell[cl]{Model compression\\\& optimization\\\cite{han2015deep,jacob2018quantization,gysel2018ristretto,fedorov2019sparse,lin2020mcunet}} & \makecell[cl]{network,\\layers,\\connections} & \makecell[cl]{post-training,\\training${^*}$}  & no & no & N/A & M7~\cite{lin2020mcunet} \\\hline
        \makecell[cl]{Sparse update \\~\cite{lin2022ondevice,profentzas2022minilearn,wang2019e2train,kwon2024tinytrain,huang2023elastic}} & \makecell[cl]{backward\\layers, \\connections}  & \makecell[cl]{pre-runtime,\\runtime} & no & \makecell[cl]{\cite{wang2019e2train}} & N/A &\makecell[cl]{M7~\cite{lin2022ondevice}\\M4~\cite{profentzas2022minilearn}} \\\hline
        \makecell[cl]{Reduced-precision\\training~\cite{zhu2016trained,li2022ternary}} & \makecell[cl]{arithmetic,\\data } & runtime  & yes/no & no & no & N/A\\\hline
        NITI~\cite{wang2022niti} & \makecell[cl]{arithmetic,\\data}  & runtime & {{yes}} & {{yes}} & no & N/A \\\hline
        Tin-Tin (\textit{this work}) & \makecell[cl]{arithmetic,\\data}  & runtime  & {{yes}} & {{yes}} & {{yes}}  & {M0, M4} \\\hline
    \end{tabular}
    \end{center}
    \footnotesize\emph{$^*$} The term ``training'' is used separate from ``runtime'' to emphasize the clear distinction between model training phase and the inference phase in these works. 
    
    \footnotesize\emph{$^{\dagger}$} ``M'' refers to the Cortex M series. 
\end{table*}
\section{Related Work}\label{sec:related}

\paragraph{Model Compression \& Optimization} Many methods have been developed to fit large ML models for deployment on small devices with limited memory, as surveyed by \citet{liang2021pruning}. Techniques include pruning~\cite{han2015deep,he2018amc,liu2019metapruning,He_2017_ICCV,liu2020autocompress,li2016pruning}, quantization~\cite{banner2019post,wang2019haq,zhou2018adaptive,gysel2018ristretto}, and neural architecture search~\cite{lin2020mcunet,fedorov2019sparse,cai2018proxylessnas,jin2019autokeras,tan2019mnasnet,liberis2021munas}.  Some methods integrate these techniques during training \cite{alvarez2016learning,wen2016learning}, such as quantization-aware training \cite{jacob2018quantization}, which simulates quantization effects, and IQN \cite{zhou2017incremental}, which incrementally quantizes models through re-training. While these methods effectively reduce model size and achieve faster inference, they do not address the memory constraints during training. In contrast, our work focuses on continual on-device learning.

\paragraph{Memory-saving Techniques} Efficient on-device training under memory constraints has been widely explored for DNNs. Some methods reduce memory usage by recomputing some results on-the-fly rather than saving them~\cite{pan2021mesa,chen2016training,gim2022memory,wang2022melon}, which incurs significant computation overhead and exacerbates on-device training time. Other works~\cite{lin2022ondevice,cai2020tinytl,profentzas2022minilearn,qu2022p,huang2023elastic,wang2019e2train,kwon2024tinytrain} explore sparse updates that select only a subset of layers (and tensors) to update during backpropagation to reduce memory consumption and computation. However, they require non-trivial training, such as the tiny training engine~\cite{lin2022ondevice} and meta-training~\cite{kwon2024tinytrain}, or extensive analysis to identify important layers and channels relevant to the target data (e.g., contribution analysis~\cite{lin2022ondevice}, pruning~\cite{profentzas2022minilearn}). This process may not be feasible if the training dataset is unavailable, {e.g., if it is confined to MCUs and cannot be easily moved to more powerful devices for analysis,} and adapting to new data distributions after deployment is challenging. Additionally, these techniques mainly target deep networks and involve extensive floating-point operations. In contrast, our work focuses on simple learning tasks common on battery-powered sensor devices dealing with time-series data (e.g., temperature, acceleration) and aims to minimize floating-point operations.

\paragraph{Reduced-precision Training}  The use of low bit-width reduced-precision data types for model parameters has received significant attention for its potential to reduce the computation cost  (e.g., binary~\cite{courbariaux2016binarized,hubara2018quantized}, ternary~\cite{zhu2016trained,li2022ternary}, XNOR~\cite{rastegari2016xnor}, WAGE~\cite{wu2018training}).  However, these solutions remain impractical for low-power sensor devices because  heavy intermediate floating-point calculations are still required at certain computation stages, which can be highly inefficient on tiny devices. For instance, WAGE~\cite{wu2018training} accumulates weights with high precision in fp32 before constraining to 8-bit integers. Some of them rely on non-standard number systems (e.g., binary, nonuniform~\cite{polino2018model}) that requires dedicated hardware support.
Several related works address the determination of quantization scales for reduced-precision training. PACT~\cite{choi2018pact} optimally determines the quantization scale for activations, LSQ~\cite{esser2019learned} and LQ-Nets~\cite{Zhang_2018_ECCV} optimize the quantizer for weights, while LSQ+~\cite{bhalgat2020lsq+} extends LSQ to asymmetric quantization. These approaches apply to full-precision models.

NITI~\cite{wang2022niti}, an integer-only training framework for DNNs, is the closest work to ours. NITI addresses the dynamic range issue  through an innovative exponentiation scaling scheme for integer values. Unlike NITI, which fixes the weight scales, our integer-based rescaling scheme allows weight values to be dynamically adjustable during the entire training process, providing additional flexibility and precision. Moreover, NITI focuses on GPU and FPGA deployments, while we design Tin-Tin for end-to-end MCU sensing applications.

\section{Background \& Motivation}\label{sec:background}
{Our work is motivated by two key factors. First, integer-based model quantization can substantially reduce device memory requirements. Second, floating-point operations consume a considerable amount of energy. We then demonstrate that model weights can vary significantly during training, highlighting the necessity for a dynamic weight range.}

\begin{figure}
    \centering
    \includegraphics[width=\linewidth]{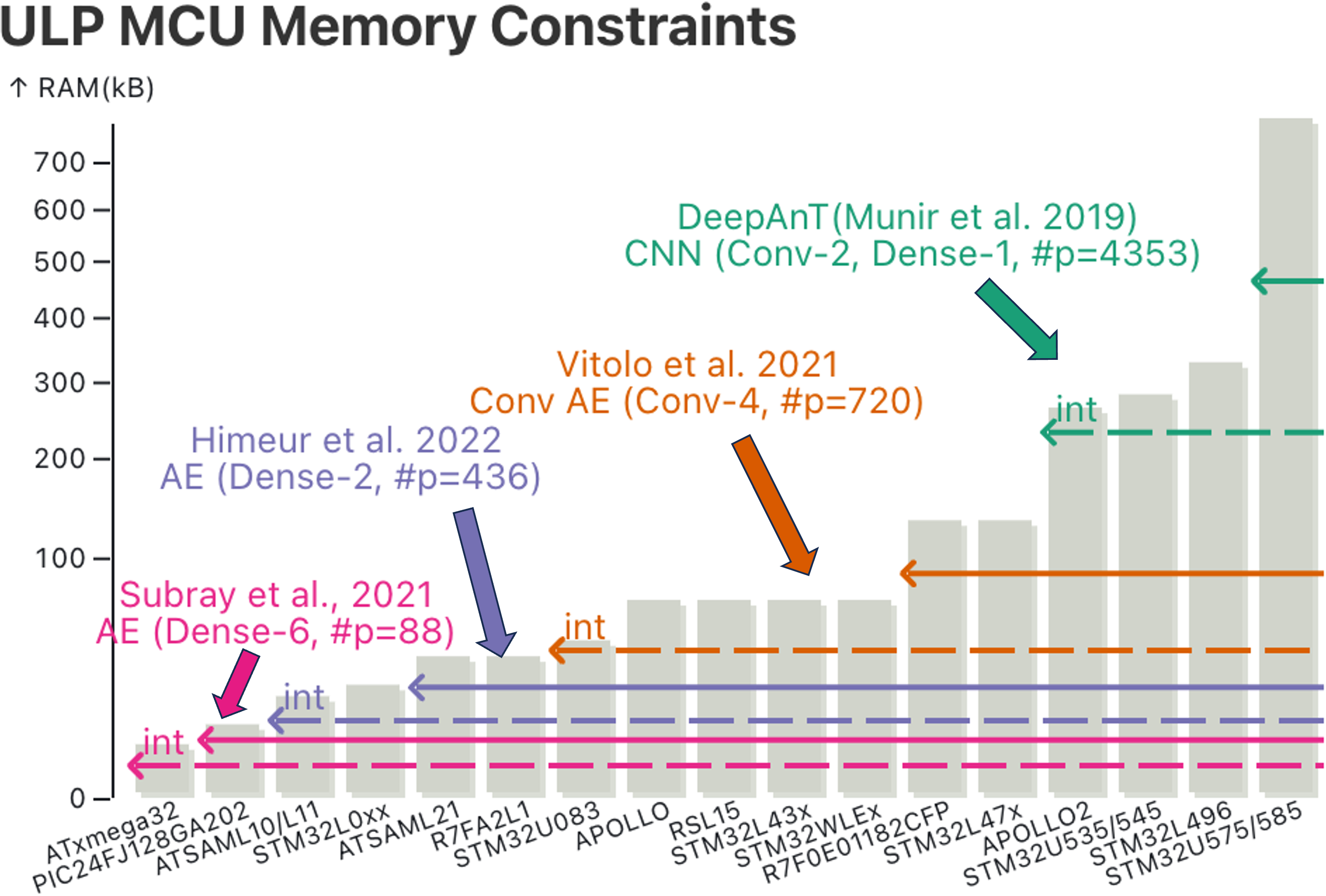}
    \caption{Data RAM of ULP MCUs and the memory requirements for various ML applications, including anomaly detection (\citet{munir2019deepant,vitolo2021lowpower,himeur2022detection}) and signal classification (~\citet{subray2021spectrum}). Solid lines: \texttt{fp32}; dashed lines: \texttt{int8}. For DeepAnT~\cite{munir2019deepant}, a kernel size of 3 and a pooling layer that preserves the output dimensions are assumed. Integer-based ML frameworks significantly reduce RAM requirements, expanding the range of devices on which ML applications can be deployed.}
    \label{fig:memory}
\end{figure}

\subsection{Device Memory Constraints}\label{sec:memory_mot}
Low-power MCUs typically have only hundreds of KB of memory (data RAM) or less to maintain low-power, low-budget operations. Figure \ref{fig:memory} shows the RAM sizes of some popular low-power devices from EEMBC ULPMark\textsuperscript{\textregistered}, an ultra-low power MCU benchmark that quantifies the low-power behavior of MCUs\footnote{https://www.eembc.org/ulpmark/}.

Limited memory budgets significantly constrain ML applications on tiny devices. For both inference and training, the model needs to be loaded into memory. Training requires substantially more memory than inference due to the additional memory needed for backpropagation. In inference-only scenarios, all parameters of the pre-trained model can be accessed in read-only mode, and intermediate activations between layers can be discarded as data flows to the next layer. However, during training, activations must be stored to calculate gradients during the backward pass, further exacerbating memory consumption.

In general, MCUs are not designed for complex computations, such as deep vision tasks, without significant hardware and energy support. Therefore, targeted tiny learning applications should be lightweight and simple, benefiting autonomous and continual monitoring without heavily burdening battery life. One motivating application is anomaly detection, achievable via a simple auto-encoder (AE). Figure \ref{fig:memory} shows the model architecture and the number of parameters (\#p) used in four works for ML-based anomaly detection~\cite{munir2019deepant,subray2021spectrum,vitolo2021lowpower,himeur2022detection}. As shown in Figure \ref{fig:forward_backward}, we estimate the memory usage considering three types of memory consumption for training: (1) loading all model parameters, (2) storing activations during the forward pass, and (3) dynamically allocated memory during the process (e.g., temporary buffers for storing gradient results). The first two are typically statically allocated on MCUs, while the third is estimated as the sum of the sizes of the largest weight and activation in full precision (\texttt{fp32}/\texttt{int32}) across layers for storing intermediate activation and gradient results. The estimated memory usage of using \texttt{fp32} and \texttt{int8} for storing model parameters and activations with a batch size of 32 is shown in Figure \ref{fig:memory}. Using integer weights and parameters can mitigate memory constraints and enable learning on more low-power devices with as little as 2 KB of memory (e.g., ATxmega32).

\subsection{Expensive Floating Point Operations}\label{sec:int-motive}
\begin{figure}
    \centering
    \includegraphics[width=\linewidth]{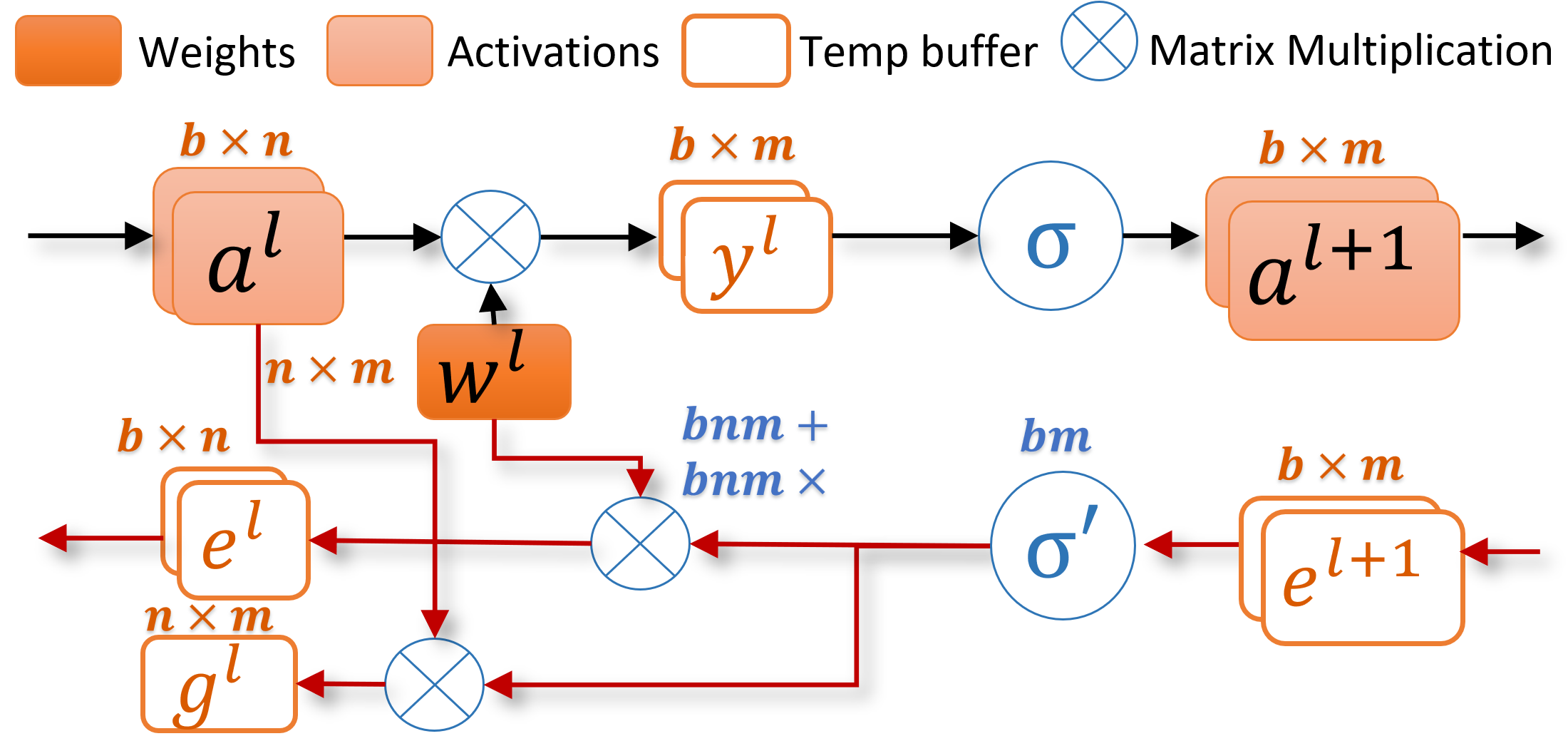}
    \caption{Dataflow illustrating the memory requirements and computation costs of the forward (black) and backward (red) passes for a linear layer with input  dimension $n$, output dimension $m$ and batch size $b$.}
    \label{fig:forward_backward}
\end{figure}
\begin{table}
    \centering
    \caption{Measured latency, power and energy consumption for a single matrix multiplication involving two 8x8 matrices. Integer (int8) computations consume significantly less energy compared to floating point (fp32) on all devices.}
    \label{tab:energy_measure}
    \begin{center}
    \begin{tabular}{l l l r r r}\toprule
         \makecell[cl]{Device} & \makecell[cl]{Freq\\ (Hz)} & & {t ($\mu$s)} & \makecell[cl]{P (mW)} & \makecell[cl]{E ($\mu$J)} \\\hline
        \multirow{2}{*}{\makecell[cl]{ Feather M0 \\ATSAMD21}} &  \multirow{2}{*}{\makecell[cl]{48M}} & \makecell[cl]{fp32} &  3550.86 & 44.15 & 156.77\\\cline{3-6}
        & & \makecell[cl]{int8} & \textbf{148.06} & 35.63& \textbf{5.28}\\\hline
        \multirow{2}{*}{\makecell[cl]{ Feather \\ RP2040}}  &  \multirow{2}{*}{\makecell[cl]{125M}} & \makecell[cl]{fp32} & 641.37 & 102.94 & 66.02 \\\cline{3-6}
        & & \makecell[cl]{int8} & \textbf{74.32}& 109.72& \textbf{8.15}\\\hline        \multirow{2}{*}{\makecell[cl]{ expLoRaBLE \\Apollo3/FPU}} &  \multirow{2}{*}{\makecell[cl]{96M}}  & \makecell[cl]{fp32} & 133.45 & 13.29 & 1.77\\\cline{3-6}
        & & \makecell[cl]{int8} & \textbf{100.68 }& 13.45& \textbf{1.35}\\\hline
    \end{tabular}
    \end{center}
\end{table}
\begin{figure*}
    \centering
    \includegraphics[width=.8\linewidth]{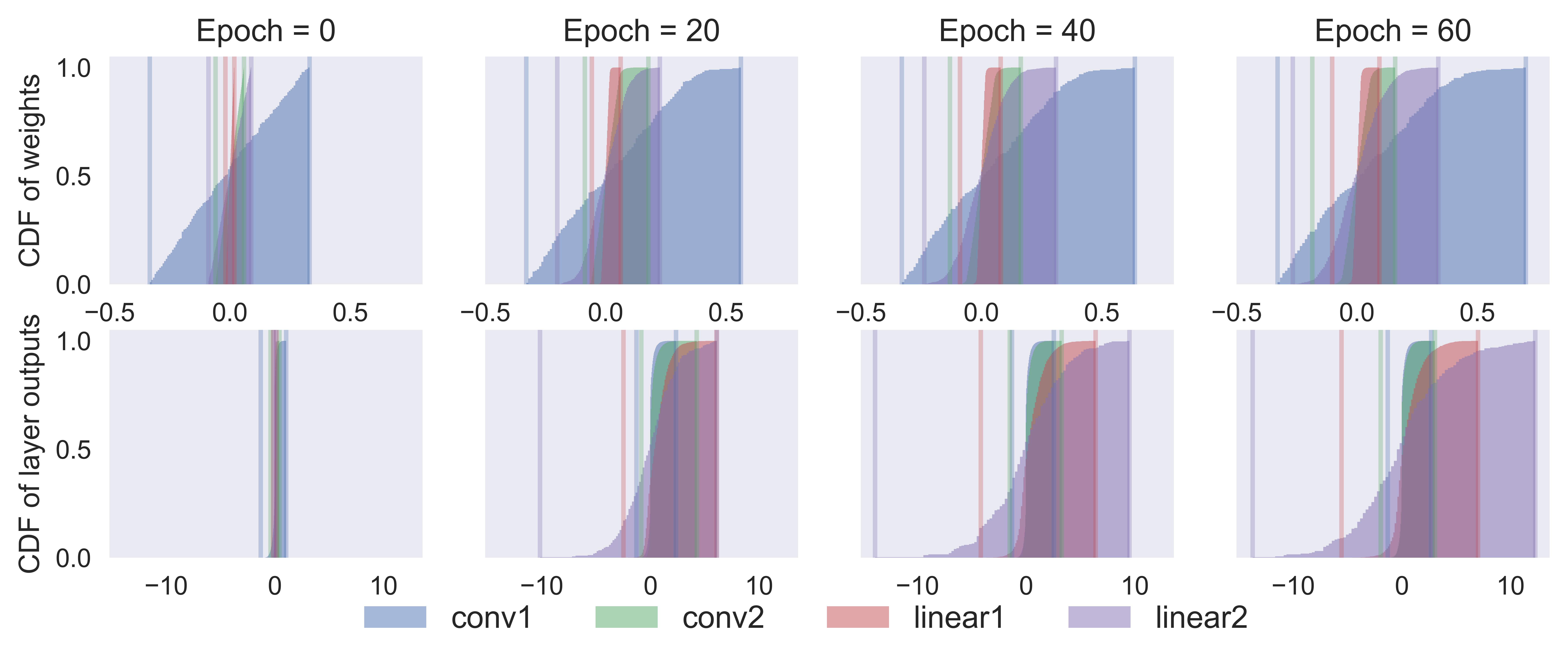}
    \caption{The value distributions of weights and layer outputs change significantly across layers and epochs during the training of a four-layer model for MNIST, demonstrating the need for dynamic representation ranges.}
    \label{fig:dr}
\end{figure*}

Training a neural network requires significant computational resources, mainly for matrix multiplication.  Figure \ref{fig:forward_backward} shows the computation involved in the forward and backward passes of a linear layer $l$, which includes one matrix multiplication in the forward pass and two matrix multiplications in the backward pass.  With input dimension $n$, output dimension $m$, and batch size $b$, each matrix multiplication requires $bnm$ multiply-add operations (MACs), totaling $2bnm$ floating point operations (FLOPs), in addition to activation functions on each output that require $bm$ FLOPs per pass.  The calculations for convolutional layers can be more complex, with the number of matrix multiplications increasing with the number of channels and the dimension of the output.

Performing matrix multiplications with floating points is very  expensive in terms of both latency and energy, especially on MCUs without FPUs, which must rely on slower software implementations. We measured the latency of a single matrix multiplication of two 8x8 matrices (using three nested for loops) on three MCU boards with different computational capabilities. A power analyzer was used to measure the current consumption. The results of using floating point data type (\texttt{fp32}$\times$\texttt{fp32} -> \texttt{fp32}) and integer data type (\texttt{int8}$\times$\texttt{int8} -> \texttt{int32}) are summarized in  Table \ref{tab:energy_measure}. Devices without FPUs (e.g., ATSAMD21, RP2040) are very slow on FLOPs, increasing energy consumption by up to 30 times. Therefore, replacing floating points with integers can significantly improve energy efficiency on these tiny devices with limited compute capabilities. Integer calculations are also faster on more powerful edge devices. For example, the same matrix multiplication takes  $228.37\mu s$ with floating points and  $203.05\mu s$ with integers on a laptop (MSI Summit E13 Flip Evo), and $196.86\mu s$ with  floating points and  $170.99\mu s$ with integers on an iPhone 15. Although many prior works have focused on quantizing neural networks to improve latency and energy efficiency, \textit{we are the first to incorporate integer calculations into training on resource-constrained MCUs}.

\subsection{Importance of Dynamic Range}\label{sec:dr_motivation}

\texttt{int8} has a limited representation range and precision. In a simple symmetric and uniform quantization scheme with a scaling factor $s$, the discrete values $N=-128,-127,...,127$ are mapped to $sN$. Values between $sN$ and $s(N+1)$ are  rounded to either $sN$ or $s(N+1)$. A smaller $s$ reduces the rounding error but also limits the representable range, causing values outside of $[-128\times s, 127\times s]$ to overflow and lead to inaccuracies.

Determining a proper scaling factor and range throughout training is crucial to balance reducing quantization error and avoiding overflow, ensuring training accuracy and stability. However, during model training, value distributions of variables (e.g., weights, activations) change dynamically.  As shown in Figure \ref{fig:dr}, these ranges can differ significantly across layers depending on the layer structure (e.g., kernel size, input/output dimension), activation functions (e.g., ReLU, sigmoid) and across variables (e.g., weights vs.  activations). They are also interdependent; for example, the layer output range is influenced by the distributions of both the layer input and weights, which evolve during training. 

Dynamic scale adjustment is necessary to handle these variations, ensuring each variable's range is appropriately scaled to minimize overflow and maintain low quantization error. NITI~\cite{wang2022niti} uses a block exponentiation scaling scheme to adjust scales dynamically but keeps weight scaling factors (determined by the weight initialization) fixed, leading to overflow as weights exceed the initial range after a few updates (as suggested in Figure \ref{fig:dr}). In contrast, we propose a new framework that supports dynamic weight range adjustments, reducing errors caused by improper scaling to improve overall performance.

\section{Tin-Tin}\label{sec:system}
\begin{figure}
    \centering
    \includegraphics[width=\linewidth]{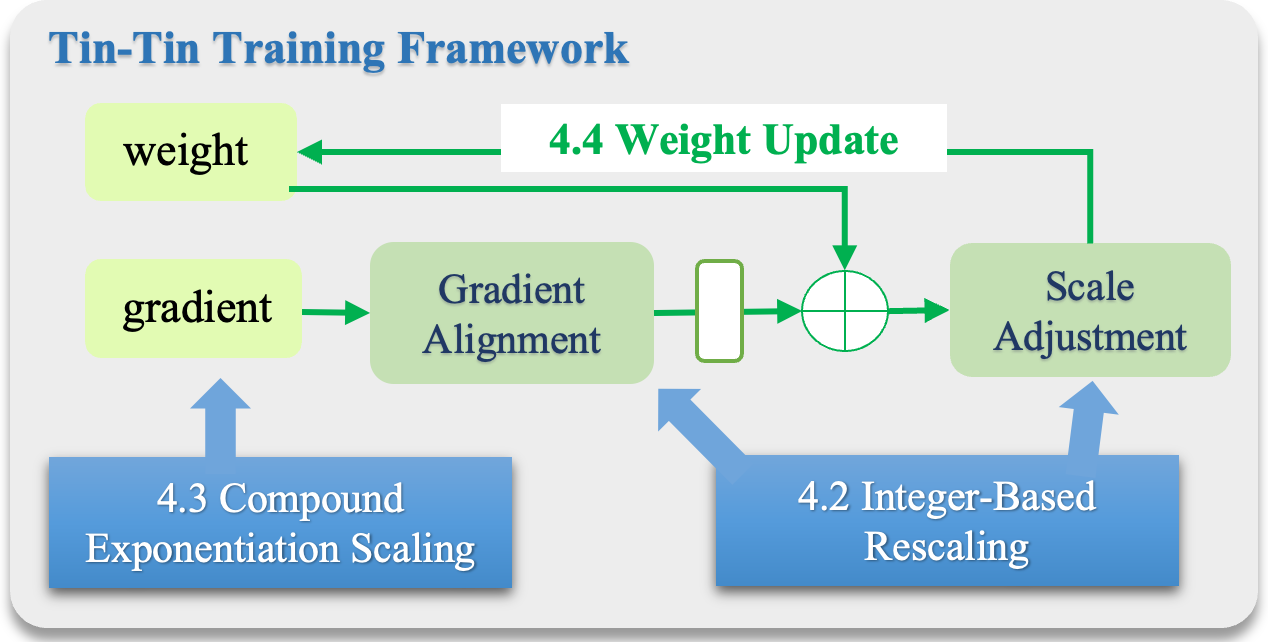}
    \caption{Tin-Tin Training Framework}
    \label{fig:tintin}
\end{figure}
We propose Tin-Tin, a framework designed for efficient integer-based neural network training on tiny devices like MCUs. As shown in Figure \ref{fig:tintin}, Tin-Tin enhances the training process by supporting dynamic scale adjustment of integer networks through a light-weight integer-based rescaling method (Sec. \ref{sec:scaleup_down}) combined with a compound exponentiation scaling scheme (Sec. \ref{sec:scaling}).  An overview of the training framework and implementation specifics is  provided in Sec. \ref{sec:overview}.  We further discuss two rescaling-enabled techniques to improve the weight updates (Sec. \ref{sec:weight_update}).
\subsection{Overview}\label{sec:overview}
\begin{figure}
    \centering
    \includegraphics[width=\linewidth]{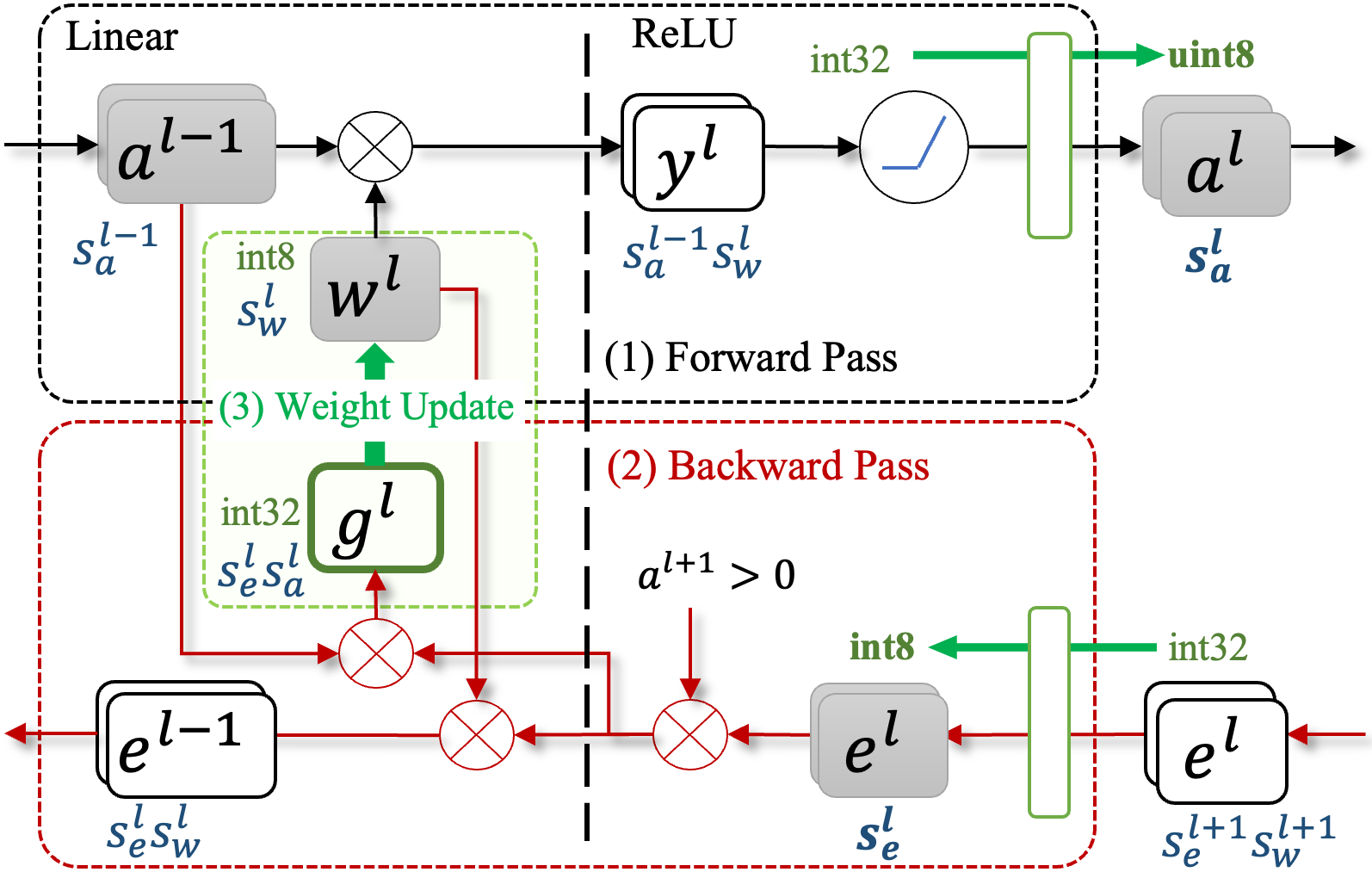}
    \caption{The forward pass (black) and backward pass (red) of a linear layer with ReLU activation are shown. 8-bit integer operands (shaded boxes) are used for all matrix multiplications.}
    \label{fig:overview}
\end{figure}
 \begin{figure*}[h]
     \centering
     \includegraphics[width=\linewidth]{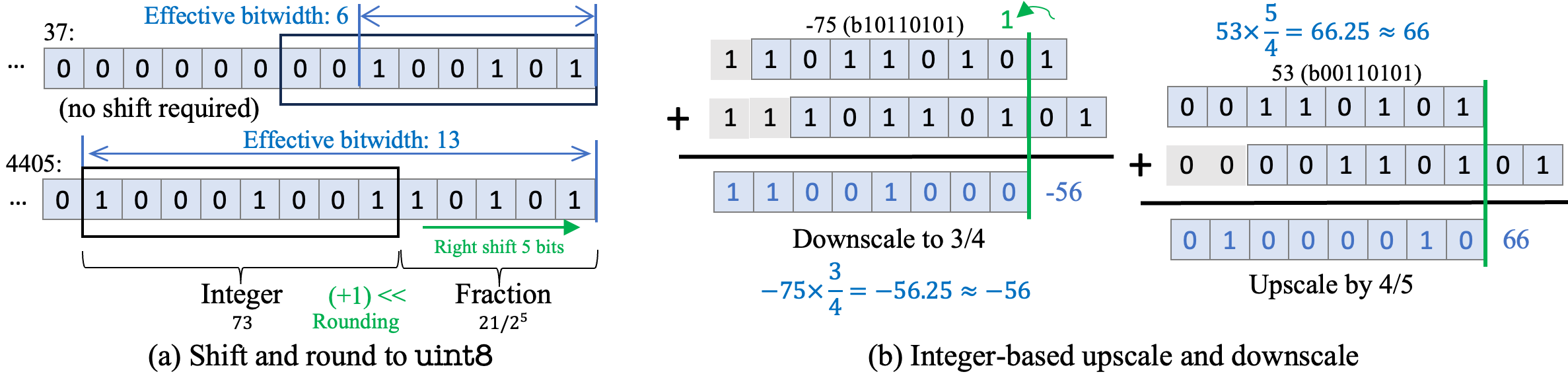}
     \caption{(a) Examples of shift-and-round to \texttt{uint8}. (b) Tin-Tin's finer-grained upscaling and downscaling, achieved through shift-and-round and \texttt{int8} addition. }
     \label{fig:shift}
 \end{figure*}
We use a linear layer with ReLU activation as an example to illustrate our approach. Figure \ref{fig:overview} shows the forward and backward passes of a layer $l$, with the layer index indicated in superscript. To replace the majority of  FLOPs with integer arithmetic, all matrix multiplications utilize 8-bit integer operands. These operands include the weights $w$,  activations $a$ from the previous layer, and backpropagated errors (i.e., gradients of the activations) $e$.  Training data $x$ that are not initially 8-bit integers are first quantized into 8-bit integers, then used as $a^0$, the input to the first layer. 

\paragraph{{Forward Pass}}
At layer $l$,  8-bit integer  activations $a^{l-1}$ from the previous layer (or quantized raw data for the first layer)  with a scaling factor $s^{l-1}_a$ are multiplied by the layer weights $w^l$ in \texttt{int8} format with a scaling factor $s^l_w$. The results of the matrix multiplication, $y_{32}^l$, are accumulated in \texttt{int32} precision with a combined scaling factor  $s^{l-1}_as^l_w$. After applying the ReLU activation, the \texttt{int32} values are rounded back to \texttt{uint8} before being passed to the next layer (details in Sec. \ref{sec:scaleup_down}).  
We use \texttt{uint8} to capture the non-negative outputs from ReLU, enhancing the precision by 1 bit than \texttt{int8}. 

\paragraph{{Backward Pass}} The backward pass propagates the error back from the final loss through the integer network.  At layer $l$,  the error $e_{32}^l$ propagated back from the next layer is initially in higher precision and is first rounded to \texttt{int8} (Sec. \ref{sec:scaleup_down}). The error is then multiplied by the derivative of ReLU, which zeros out error terms where the original activations are negative.  This adjusted error is multiplied by the layer input $a^{l-1}$ to compute the gradient $g_{32}^l$, and separately multiplied by the weight $w^l$ to derive the error $e_{32}^{l-1}$ to send to the the preceding layer. Both results are accumulated in \texttt{int32} precision, with scaling factors $s^l_es^l_a$ and $s^l_es^l_w$, respectively. 

\paragraph{{Weight Update}} Gradient descent optimizes the model parameters by following the gradient path. Traditional gradient descent updates the weight parameters by $w^{l}\leftarrow w^l - \eta /b\times g^l$, where $\eta$ is the learning rate and $b$ is the batch size. However, this scaled update step is in floating-point and much smaller than the weight scale. To apply \texttt{int32} gradients with a scaling factor of  $s^l_es^l_a$ to the \texttt{int8} weight $w^l$ with a scaling factor $s^l_w$, 
we perform integer weight updates (Sec. \ref{sec:weight_update})
\begin{equation*}
    w^{l} \gets w^{l} - INT(g^l).
\end{equation*}
Weight updates can change the value distribution of the weight, potentially causing overflow and unwanted clipping when the resulting weight values exceed 127  (\texttt{INT8\_MAX}). One of our contributions (Table~\ref{tab:related_compare}) is to allow for dynamic weight scales through integer-based scaling (Sec. \ref{sec:scaleup_down}) to improve the accuracy throughout the training.

\subsection{Integer-based Upscale and Downscale}\label{sec:scaleup_down} 
For integer-based training, it is sometimes necessary to scale the weights and intermediate results to avoid overflow when values exceed the initial quantization range. Directly multiplying integer values with a non-integer scale results in floating points and introduces FLOPs.

Simple multiplication or division by powers of two can be achieved through shift-and-round operations. This method can be used for bitwidth reduction, such as rounding \texttt{int32} to \texttt{int8} or \texttt{uint8}.  Figure \ref{fig:shift}(a) illustrates this process with two examples. Given a vector of \texttt{int32} values $V$ (e.g., $y_{32}^l$), the effective bitwidth $b^*$ is first determined, denoting the minimum number of bits required to represent the maximum absolute value $|V|$, calculated as $b^*=\ceil{\log_2(|V|)}$.  If $b^*$ is smaller than the target bitwidth $b$ (8  for \texttt{uint8} and 7 for \texttt{int8}), the \texttt{int32} values can be safely truncated to 8-bit (Figure \ref{fig:shift}(a) top). Otherwise, the \texttt{int32} values are right-shifted by $b^*-b$ to retain the most significant part of the values, with the shifted-out portion treated as the fraction optionally rounded to 1 (Figure \ref{fig:shift}(a) bottom). 

This right shifting effectively divides the original values by $2^{b^*-b}$. Similarly, integers can be multiplied by powers of two through left-shifting, which does not introduce any rounding error.  However, shift-and-round can only handle rescaling to powers of two and is inefficient for finer adjustments (e.g., scaling up by 1.2 or scaling down by 0.8). To address this, we propose an efficient integer-based rescaling scheme that utilizes shift-and-round operations and \texttt{int8} addition for scaling to non-power-of-two factors.

 We drew inspiration from the fact that any positive values can be approximated as the sums of distinct powers of 2 and scaling by powers of 2 can be efficiently achieved through shifting and {rounding}. For example, to multiply the quantized integer representation by  $5/6\approx 2^{-1}+2^{-2}+2^{-4}$, instead of converting to floating points, we approximate the result by summing the integer values shifted by  1, 2, and 4 bits (i.e., $Q$>>1+$Q$>>2+$Q$>>4). Algorithm \ref{alg:scaling}  outlines a binary decomposition process to determine the number of bits to shift given a scale factor $r$.   This method enables fine-grained scaling using only shift-and-round operations and \texttt{int8} addition, and can be directly applied to integers without FLOPs.  This integer-based scaling method can be  particularly useful in systems  where floating-point operations are too expensive or infeasible, allowing for finer-grained  adjustments to weight and activation magnitudes without the computational overhead. 

 \begin{algorithm}
    \caption{Integer-based Scaling}
    \label{alg:scaling}
    \begin{algorithmic}[1] 
    \Require {Integer $V$, scale $r>0$}
    \State $C\gets [],e\gets \floor{\log_2{r}}$
    \State $c\gets 2^e$
    \While{len($C$)$<n$ and $r>0$} \Comment{\eve{\small{limit \# of decomposition}}}
    \If{$r\ge c$}
    \State $C$.append($e$)
    \State $r\gets r-c$
    \EndIf
    \State $e,c\gets e - 1, c/2$
    \EndWhile
    \Statex \eve{\# {$e>0$: left shift, $e<0$: right shift}}
    \State $V'\gets \sum_{e\in C}\texttt{shift\_and\_round}(V,-e)$
    \end{algorithmic}
\end{algorithm}

\subsection{Compound Exponentiation Scaling}\label{sec:scaling}
In addition to upscaling and downscaling, which allows us to adjust scaling factors with integer operations, tracking scaling factors between quantized values and their corresponding real values throughout training is crucial. For example, when multiplying two values, the product carries a scaling factor that is the product of the two operands' scaling factors (i.e., $s_1Q_1\times s_2Q_2=(s_1s_2)(Q_1Q_2)$).  Additionally, complex activation functions output differently based on the scaling factors of their inputs.

Scaling operations, such as reducing from \texttt{int32} to \texttt{int8}, are intended to adjust bitwidth or the representable quantized range without altering the real values they represent. To maintain accuracy, the scaling factor must be updated accordingly. \textit {When rescaling the quantized integer representation  $Q$  by a factor $r$, the scaling factor $s$ must be adjusted by   $1/r$ to keep the real value $sQ$ constant. } Shifting operations can be tracked using powers of two, 
 but it is not enough for our dynamic scaling scheme  (Sec. \ref{sec:scaleup_down}) that allows scaling to an arbitrary factor.

To address this, we introduce a compound exponentiation scaling scheme to track the scales of activations $a^l$, weights $w^l$, gradients $g^l$, and errors $e^l$ for each layer $l$ of the model during training. We use fixed upscaling and downscaling operations, as illustrated in Figure \ref{fig:shift}(b). The fixed upscaling ratio is  $u=4/3$ with 3/4$Q$=$Q$>>1+$Q$>>2,  and the downscaling ratio is  $r=4/5$ with 5/4$Q$=$Q$+$Q$>>2.  

In Tin-Tin, any scaling operations  intended to adjust the scaling factors are achieved through a combination of these fixed operations and shift-and-round. Our scaling scheme employs three \texttt{int8} scaling exponents: a shift exponent: a shift exponent $S$ (with a weight of $2^S$ for shifting operations), and a pair of upscale and downscale exponents, $U$ and $D$, for upscaling and downscaling. The combined scaling factor using these exponents is given by:
\begin{equation*}
    s = 2^Su^Ur^D.
\end{equation*}
Each 1-bit right-shifting, upscaling, and downscaling operation increments the respective exponent, $S$, $U$, and $D$, by 1. Although limiting the operations to fixed upscaling and downscaling introduces some constraints, this compound scaling scheme achieves granular rescaling, providing 12 possible scales between 0 and 1 and 8 scales between 1 and 2 within two operations of upscaling and downscaling combined with 1-bit shifting. 

The memory requirement is only 3 bytes per variable. Additionally, the exponents for the product of a multiplication are simply the sum of the corresponding exponents of the operands: $s_1s_2 = 2^{S_1+S_2}u^{U_1+U_2}r^{D_1+D_2}$. Algorithm \ref{alg:linear_forward} shows how the scaling exponents (i.e., $S,U$ and $D$) are calculated (line 3) and propagated from one layer to the next (line 7).

\begin{algorithm}
    \caption{Linear Layer $l$ with ReLU - Forward}
    \label{alg:linear_forward}
    \begin{algorithmic}[1] 
    \Require{$a^{l-1}$, $S_a$, $U_a$, $D_a$}
            \State {$w^{l}$, $S_w$, $U_w$, $D_w$}
    \State $y^l_{32}\gets \text{ReLU}(\texttt{int\_mat\_mul}(a^{l-1},w^l))$
    \State $S, U, D\gets S_a+S_w,  U_a+U_w, D_a+D_w$
    \Statex  \color{blue}\# Shift and Round \color{black}
    \State $sr\gets \text{max}(\texttt{effective\_bitwidth}(y^l)-8,0)$
    \State $a^l\gets \texttt{right\_shift\_and\_round}(y^l_{32},sr)$
    \State $S\gets S + sr$
    \State \textbf{return} ($a^l, S, U, D$) to the next layer
    \end{algorithmic}
\end{algorithm}



\subsection{Efficient and Adaptive Update}\label{sec:weight_update}
Updating an integer network is challenging because the smallest update step is 1. This section discusses how our lightweight integer scaling techniques are applied to weight updates through gradient alignment and scale adjustment to enhance training performance.

\paragraph{Gradient Alignment} During backpropagation, a 32-bit gradient $g_{32}$ with a scaling factor $s_g$ is computed, and the goal is to update the \texttt{int8} weight $w$ with a scaling factor $s_w$. The challenge is twofold: aligning the gradient and weight scales, and determining a proper update step size.

In Tin-Tin, with our compound scaling exponents, aligning the scales is achieved via a sequence of shifting, upscaling, and downscaling operations. While left-shifting can be canceled out by right-shifting the same number of bits, upscaling and downscaling are not reversible because  $u^Ur^D\ne 1$ when $U+D>0$. Therefore, instead of aligning the gradient scale to the weight scale or vice versa, we align both the weight and gradient to a common scale $s$ that is reachable for both. This process is illustrated in Algorithm \ref{alg:update}, lines 1-6.  We select the smaller shifting exponent between the two because left-shifting, which reduces the shifting exponent, introduces no rounding error compared to right-shifting. The tailing zeros after left-shifting can also reduce rounding errors of the following upscaling and downscaling.

To determine the proper update step size, we draw on the rationale from LARS~\cite{you2017large}, which suggests that setting a layer-wise learning rate that maintains a certain ratio of the weights and gradients across layers improves training performance. Therefore, we determine the gradient updates at the scale of the weight by setting a fixed bitwidth difference, an update factor $m$, between the weight and the gradient.  After aligning the weight and gradient scales, we find the effective bitwidth $b_w^*$ of the weight and then shift and round the gradient to have an effective bitwidth of $b_w^*-m$, where $m<7$ (Algorithm \ref{alg:update}, lines 9-12). This ensures the absolute value of the gradient updates is within  $2^{-m}$ of the weight range, then we subtract the shifted value from the weight (line 13). This method has proven effective in~\cite{wang2022niti}. 

\paragraph{Scale Adjustment} Our proposed gradient alignment method preserves most of the gradient information by aligning both the gradient and the weight to a higher precision scale, avoiding direct truncation to low-bitwidth as done in~\cite{wang2022niti}. This approach prevents \textit{unwanted} weight overflow during updates. Since the weight is now in high precision, it needs to be rounded back to 8-bit precision. We achieve this using our integer-based rescaling method described in Section~\ref{sec:scaleup_down}. 

Given the current weight scale $s$ and a target scale $t$, the goal is to find a sequence of upscaling, downscaling, and shifting operations that adjust the scale to $t$.  This can be efficiently achieved by various combinations of our scaling operations. To facilitate this process, we pre-compute a list of scales between 0.5 and 2 that can be reached within two upscaling or downscaling operations (e.g., 0.64=$r^2$,  achievable through two downscaling operations). We then scale down the weights by combining one of these predetermined rescale ratios with shifting operations, effectively rounding the weight back to 8-bit precision. 

Our rescaling scheme offers a lightweight solution for rescaling in an integer network. However, improper use of rescaling can cause the value range to expand excessively, leading to poor learning performance. In Tin-Tin, we incorporate clipping techniques and maintain proper scaling factors among values in the network. This helps keep values within a reasonable range and maintain appropriate norms from layer to layer, which generally leads to good convergence.~\cite{glorot2010understanding}.

\begin{algorithm}
    \caption{Weight Update}
    \label{alg:update}
    \begin{algorithmic}[1] 
    \Require {$w^{l}$, $S_w$, $U_w$, $D_w$; $g_{32}^{l}$, $S_g$, $U_g$, $D_g$}
    \Procedure{Scale Alignment}{$Q_1$,$S_1$,$U_1$,$D_1$,$Q_2$,$S_2$,$U_2$,$D_2$}
    \State $S,U,D\gets \min(S_1,S_2),\max(U_1,U_2),\max(D_1,D_2)$
    \State $\texttt{left\_shift}(Q_1,S_1-S)$ or $\texttt{left\_shift}(Q_2,S_2-S)$
    \State $\texttt{upscale}(Q_1,U-U_1)$ or $\texttt{upscale}(Q_2,U-U_2)$
    \State $\texttt{downscale}(Q_1,D-D_1)$ or $\texttt{downscale}(Q_2,D-D_2)$
    \EndProcedure
    \State $w',g'\gets \texttt{scale\_alignment}(w^l,S_w, U_w, D_w, g_{32}^{l}, S_g, U_g, D_g)$
    \State $b_g^*\gets \texttt{effective\_bitwidth}(g')$
    \State $b_w^*\gets \texttt{effective\_bitwidth}(w')$
    \State $b\gets b_w^* - m$
    \State $g'\gets \texttt{shift\_and\_round}({g'}, b_g^*-b)$
    \State $w' \gets w' - g'$ \Comment{\eve{Apply updates}}
    \State $w'\gets \texttt{rescale}(w')$
    \State $b_w \gets \texttt{effective\_bitwidth}(w')$
    \State $w^l\gets \texttt{shift\_and\_round}(w', b_w-7)$
    \end{algorithmic}
\end{algorithm}

\section{Evaluation}\label{sec:eval}

We evaluate Tin-Tin through two case studies of on-device training (Section~\ref{sec:case-studies}), as well as simulations on the standard MNIST image classification dataset (Section~\ref{sec:exp_rescale}). 
\begin{figure}[t]
    \centering
    \includegraphics[width=\linewidth]{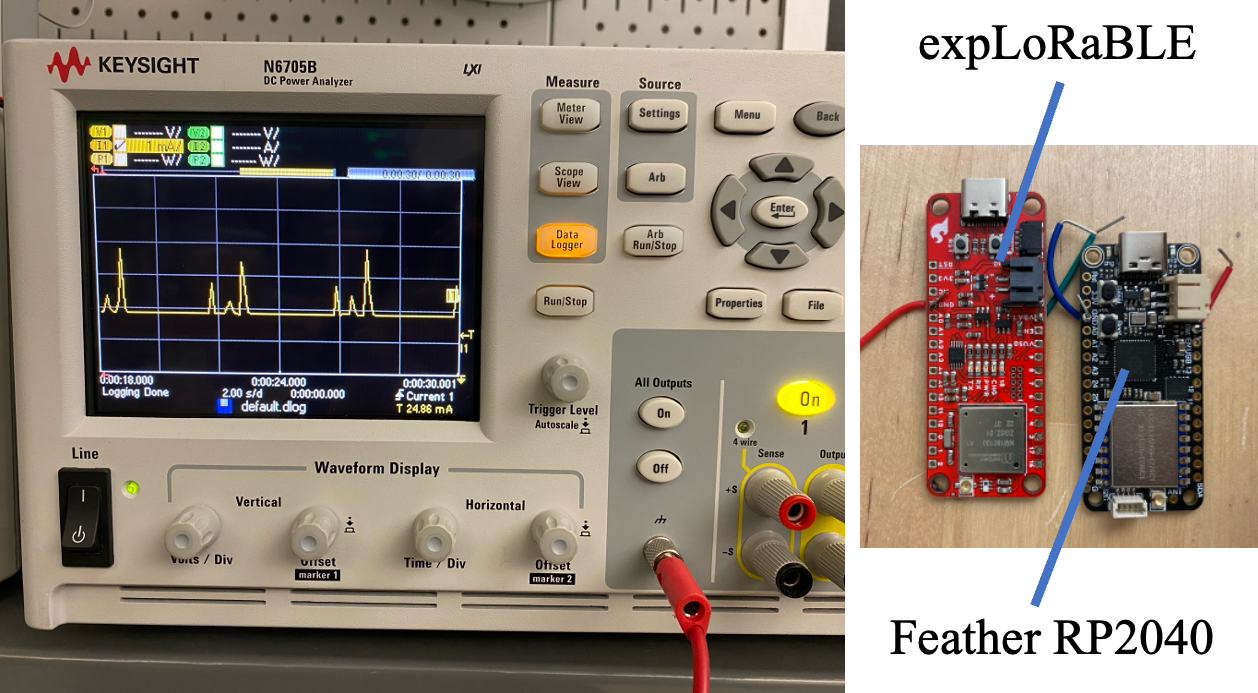}
    \caption{DC power analyzer used for power measurement and MCUs for evaluation. }
    \label{fig:exp_setup}
\end{figure}
\subsection{On-Device Case Studies}~\label{sec:case-studies}
We conduct case studies and evaluate our framework on MCUs with a focus on energy and memory efficiency. The evaluation uses two devices: (1) Adafruit Feather RP2040, featuring a Raspberry Pi Pico with a Cortex M0+ dual core and 264 KB RAM, and (2) SparkFun LoRa Things Plus (expLoRaBLE), which uses the Ambiq Apollo3 ARM Cortex M4 with FPU and 384 KB RAM. We consider the following sensor-based learning applications:

\paragraph{\textbf{Motor Bearing Fault Detection}} We consider a scenario for automatic motor bearing fault detection using vibration sensors. The sensor trains an autoencoder locally on its data. To evaluate the training process, we use an extract from the CWRU bearing dataset~\cite{cwru}, containing electric motor bearing vibration data sampled at 12 kHz in both normal and faulty states. Specifically, the fan end (FE) accelerometer data at 1797 rpm is used. A window of 32 is used as inputs. A 4-layer dense network with 24 hidden nodes and ReLU activation serves as an autoencoder. The model is trained on normal baseline data using MSE (mean-squared error) loss to learn normal conditions. During testing, the MSE (reconstruction loss) is measured on both normal and faulty conditions to evaluate the model's performance.

\paragraph{\textbf{Spectrum Sensing}} We implement a spectrum sensing scenario for radio signal classification. Following~\cite{subray2021spectrum}, we use an autoencoder with five dense layers for on-device radio signal classification. The dataset~\cite{subray2021spectrum} includes I/Q samples of LTE and WiFi signals. The autoencoder is trained on LTE signals to distinguish between a mix of WiFi and LTE signals. The network architecture includes 3 hidden layers of 32, 16 and 32 nodes, using ReLU activations and MSE loss to learn LTE signals. The input comprises a history window of 16, each including real and imaginary components, phase, and amplitude values from I/Q samples, flattened into an array of 64. The model's performance is evaluated based on its ability to distinguish between LTE and WiFi signals.

\paragraph{Setup} The MCUs are programmed through Arduino. For both applications, we implement the training of autoencoders on the MCUs and compare the training efficiency using full precision floating points versus our proposed Tin-Tin method. We measure latency, memory usage, energy consumption, and record training and validation loss.

Latency for forward and backward passes is measured separately using on-board clocks and averaged over 100 training iterations. Energy measurements are conducted using a Keysight DC Analyzer N6705B (Figure \ref{fig:exp_setup}). Since this equipment only supports sampling at 10 Hz, which is insufficient to capture all transient current behaviors, we divide the computation into forward pass, backward pass, and loss calculation. For each part, we measure the average current consumption by running the computations in a loop for 30 seconds. The measured average current is then multiplied by the measured time for each part of the computation, and the total energy costs are summed up. Each training experiment is repeated five times to obtain the average training loss.


\begin{figure*}
    \centering
    \includegraphics[width=\linewidth]{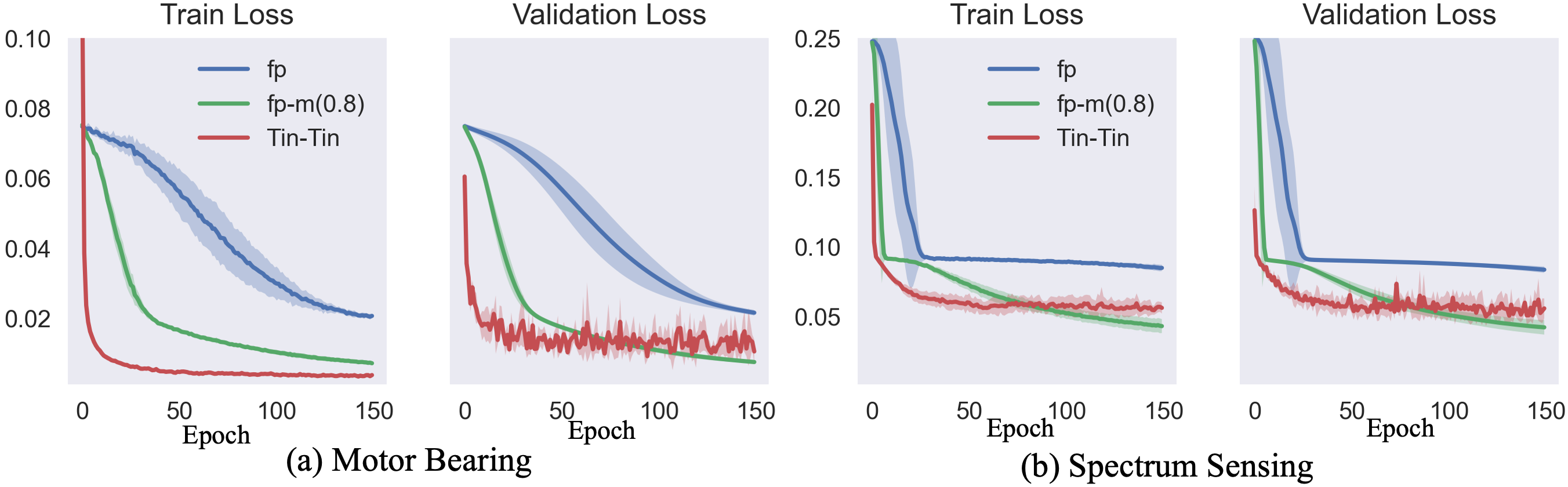}
    \caption{Training and validation MSE loss for two case study applications: (a) motor bearing fault detection and (b) spectrum signal classification. Tin-Tin rapidly reduces training loss at the beginning compared to the floating point methods. }
    \label{fig:case_study}
\end{figure*}

\subsubsection{ Performance} 
We evaluate the training efficiency and model performance after training on MCUs. For the traditional floating point implementation, a fixed learning rate of 0.1 with and without 0.8 momentum is used. {We use fp-m to denote the floating point implementation with momentum.} The training and validation losses, averaged across five experiments with a batch size of 128, are shown in Figure \ref{fig:case_study}. The fixed learning rate without momentum converges the slowest in both scenarios and is prone to getting stuck in local optima.  The use of momentum helps mitigate this issue. Tin-Tin  rapidly reduces the training loss at the beginning of the training, efficiently decreasing the number of training rounds required to achieve a desired model performance.

The reconstruction loss (MSE) on unseen test data, including both normal and abnormal data (faulty motor bearings and non-LTE signals in spectrum sensing), is summarized in Table \ref{tab:e1}. A good autoencoder typically yields low loss on normal data and high loss on anomaly data. The test results indicate that fixed learning rate floating point methods are inefficient in achieving convergence in both cases. Although momentum helps to some extent, it fails to adequately detect faulty states in the motor bearing scenario, as shown by similar reconstruction errors on normal and abnormal data. In contrast, Tin-Tin enables the model to converge much faster, maintaining acceptable performance within a manageable error bound in both cases.

Additionally, we experimented with adding a check for zero values during matrix multiplication to skip MAC operations when one of the operands is zero.  The average percentages of skipped MACs in the forward and backward passes are summarized in Table \ref{tab:e1}, suggesting that leveraging sparsity can significantly  reduce computation costs. 

\subsubsection{Resource Usage}We compared the resource efficiency of Tin-Tin and traditional floating point implementations in terms of memory and energy consumption on RP2040 and expLoRaBLE for different batch sizes, with the results summarized in Table  \ref{tab:e2}. Generally, the latency of the forward and backward passes  is proportional to the batch size, as the batch of data can only be handled sequentially on MCUs. The energy consumption is almost linear with the total computation time. On devices like RP2040 without FPUs, Tin-Tin significantly reduces the training time required for each iteration by replacing FLOPs with integer arithmetic, thereby reducing computation time and saving energy by up to 83\%. Tin-Tin also effectively reduces memory requirements for on-board training by using 8-bit integers instead of 32-bit floating points, reducing memory load by up to 40\%.  

However, Tin-Tin runs slower than floating points on the FPU-enabled expLoRaBLE, where FLOPs can be efficiently handled by FPUs within a few clock cycles. Further investigation revealed that the compiled assembly code for integer arithmetic is not optimized on expLoRaBLE. This demonstrates that integer-based training does not guarantee more efficiency than floating points under all circumstances{, though Tin-Tin still reduces the memory requirements significantly compared to floating point deployments}. Future development could benefit from system-algorithm co-design while aiming for generalizability across MCU platforms.

\begin{table}
    \centering
    \caption{Model performance and computational saving. Tin-Tin (int) requires significantly fewer MACs than floating point implementations, but achieves comparable or better reconstruction MSEs (lower is better on normal data, and higher is better on anomalous data).}
    \label{tab:e1}
    \begin{center}
    \begin{tabular}{l l c c c c}\toprule
    &  & \multicolumn{2}{c}{Reconstruction MSE} & \multicolumn{2}{c}{MAC Reduced} \\\cmidrule(lr){3-4}\cmidrule(lr){5-6}
    &  & \makecell[c]{Normal} & \makecell[c]{Anomaly} & \makecell[c]{Fwd} & \makecell[c]{Bwd} \\\hline
    \multirow{2}{*}{\makecell[cl]{Motor}} & \makecell[c]{fp} & 0.0192 & 0.0070 & - & -  \\
& \makecell[c]{fp-m} & 0.0073 & 0.0065 & - & -  \\
     & \makecell[c]{int} & \textbf{0.0050} & \textbf{0.0107}  & 79\%  &  30.7\%\\\bottomrule
    \multirow{2}{*}{\makecell[cl]{Spectrum}} &  \makecell[c]{fp} &  0.086 & 0.124 & - & -  \\
    & \makecell[c]{fp-m} & 0.038 & 0.082 & - & -  \\
     & \makecell[c]{int} & 0.050 & 0.095 & 86\%  &  58\%\\\bottomrule
    \end{tabular}
    \end{center}
\end{table}

\begin{table*}
    \centering
    \caption{Measured average latency, energy consumption, and memory usage per training iteration for different batch sizes (BS) on RP2040 and expLoRaBLE. Tin-Tin (int) significantly reduces memory requirements, and on the Feather RP2040 significantly reduces energy requirements, compared to floating point implementations (fp) for both case studies.}
    \label{tab:e2}
    \begin{center}
    \begin{tabular}{c l l r r r r r r r r}\toprule
   & &  & \multicolumn{4}{c}{Feather RP2040} & \multicolumn{4}{c}{expLoRaBLE w/ FPU} \\\cmidrule(lr){4-7}\cmidrule(lr){8-11}
        & \makecell[cl]{BS} &  & \makecell[cl]{Forward\\Time (ms)} & \makecell[cl]{ Backward \\ Time (ms)} & \makecell[c]{Energy \\ (mJ)} & \makecell[c]{Memory \\ (KB)}   & \makecell[cl]{Forward\\Time (ms)} & \makecell[cl]{ Backward \\ Time (ms)}  & \makecell[c]{Energy \\ (mJ)} & \makecell[c]{Memory \\ (KB)}\\\hline
        
\multirow{6}{*}{\makecell[c]{Motor\\Fault\\Detection }} & \multirow{2}{*}{\makecell[cl]{16}} & \makecell[cl]{fp} &56.46 &114.02&18.51  & 30.27  &  13.67 & 27.74 & 0.60 & 45.96  \\
&&\makecell[cl]{int} &  {11.26} &  {20.84} & \note{81}{3.43} &\note{24}{22.93}    &  15.75 & 31.86  & 0.69 & \note{11}{40.73}  \\\cline{2-11}
      
&\multirow{2}{*}{\makecell[cl]{32}}  & \makecell[cl]{fp} & 112.84 &224.12  & 36.59  & 40.00 &27.55 & 54.25 & 1.18 & 53.64 \\
&   & \makecell[cl]{int} &  {22.42} &  {39.98} & \note{82}{6.67} & \note{27}{29.01}  & 31.01  &  60.44 & 1.32 &\note{13}{46.74}  \\\cline{2-11}    
      
    &    \multirow{2}{*}{\makecell[cl]{64}}  & \makecell[cl]{fp} & 225.63 & 443.84 & 72.71&  59.46 &  55.33  & 114.47 & 2.45 & 77.19  \\
     &    & \makecell[cl]{int} & {44.97} &  {77.08} & \note{82}{13.02} & \note{31}{41.04} & 62.06 & 117.60 & 2.59 & \note{24}{58.78} \\\cline{2-11}    
     
  &\multirow{2}{*}{\makecell[cl]{128}}  & \makecell[cl]{fp} & 451.27 & 883.49 &  144.96 & 98.37 & 112.77 & 244.34 &5.15  & 116.10 \\
      &   & \makecell[cl]{int} &  {89.53} &  {151.97} & \note{82}{25.80} &  \note{34}{65.10} &  131.75 & 254.53  & 5.54 & \note{29}{82.84}\\\hline

\multirow{6}{*}{\makecell[c]{Spectrum\\Sensing}} & \multirow{2}{*}{\makecell[cl]{16}} & \makecell[cl]{fp} & 115.48 & 232.02 & 37.13 & 47.86 &  27.86 & 56.86 &1.22 &  65.67 \\
&  & \makecell[cl]{int} & {21.92} & {44.90} & \note{80}{7.26}  &\note{31}{33.04}  & 30.40 & 68.33 & 1.42 & \note{23}{50.78} \\\cline{2-11}

&\multirow{2}{*}{\makecell[cl]{32}}  & \makecell[cl]{fp} & 230.89 & 456.56 & 73.45 & 64.58 & 55.69 & 118.65 & 2.51 & 82.31 \\
&  & \makecell[cl]{int} & {43.64} & {85.16} & \note{81}{13.99} & \note{34}{42.32}  & 60.22   & 127.78 & 2.71& \note{27}{60.06} \\\cline{2-11} 
      
&    \multirow{2}{*}{\makecell[cl]{64}}  & \makecell[cl]{fp} & 461.70 & 905.63 & 146.09 & 97.86 & 111.71 & 255.82 & 5.30 & 115.59 \\
 &    & \makecell[cl]{int} & {86.72} & {164.02} & \note{81}{27.23} &\note{38}{60.88} & 126.64 & 272.45 & 5.75& \note{32}{78.62}  \\\cline{2-11} 
     
 &    \multirow{2}{*}{\makecell[cl]{128}}  & \makecell[cl]{fp} & 923.33 & 1803.77 & 291.38 & 164.42 & 237.75 & 513.88  & 10.84 & 182.15  \\
     &    & \makecell[cl]{int} & {174.06} & {322.10} & \note{82}{53.89} & \note{40}{98.00} &  252.45 & 540.79 & 11.44& \note{36}{115.74}  \\\hline 
    \end{tabular}
    \end{center}
\end{table*}
\begin{figure*}
    \centering
    \includegraphics[width=\linewidth]{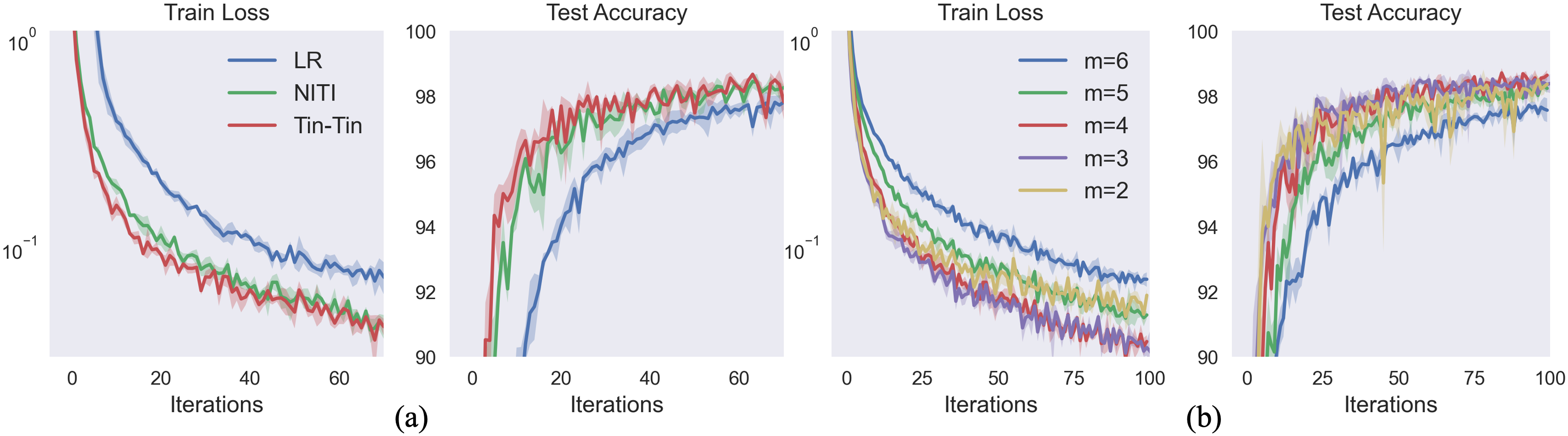}
    \caption{Evaluation on MNIST dataset. (a) Training loss and test accuracy using LeNet-5, comparing Tin-Tin with NITI and traditional learning rate (LR) updates.  (b) The impact of update factors $m$ on Tin-Tin's integer updates.}
    \label{fig:eval}
\end{figure*}

\begin{table}
    \centering
    \caption{Test accuracy on MNIST after 250 iterations, with each iteration training on 15 batches of 256 data points. }
    \label{tab:mnist}
    \begin{center}
    \begin{tabular}{c c c c c c}\toprule
{NITI} &   \multicolumn{5}{c}{Tin-Tin}  \\\cmidrule(lr){1-1}\cmidrule(lr){2-6}
 m=4   & m=2 & m=3 & m=4 & m=5 & m=6 \\
99.2\%&  96.84\% & 97.01\% & 99.2\% & 98.98\% & 98.69\%
    \\\bottomrule
    \end{tabular}
    \end{center}
\end{table}

\subsection{MNIST Dataset}\label{sec:exp_rescale}
We further evaluate Tin-Tin's rescaling on the standard benchmark  MNIST dataset using  the LeNet-5 model. 

\subsubsection{Rescaling Effect} We compare Tin-Tin with NITI and floating-point updates, using a learning rate of 0.01 and a momentum of 0.8. Both Tin-Tin and NITI maintain the most significant few bits of the gradients for weight updates.  In this experiment, we set $m$, the difference between the most significant bit of weights and gradients (Sec. \ref{sec:weight_update}), to 4 for both Tin-Tin and NITI for a fair comparison.    Figure \ref{fig:eval} (a) shows the training loss and test accuracy averaged over five experiments when training LeNet-5 on MNIST.  Both NITI and Tin-Tin, using integer updates, converge much faster than using a standard learning rate, with Tin-Tin slightly outperforming NITI. This improvement is attributed to the initial weight upscaling in Tin-Tin, resulting in larger-scale updates early in the training process, which accelerates learning compared to the weight-fixed scale in NITI.

\subsubsection{Update Factor} The update factor $m$ is an importance hyperparameter in Tin-Tin, as it determines the size of the weight updates. A larger $m$ increases the scale differences between the weights and updates, resulting in smaller update steps. We train on the MNIST dataset with different update factor $m$,  and the results are shown in Figure \ref{fig:eval} (b) and final test accuracy in Table \ref{tab:mnist}. Similar to adjusting the learning rate, $m$ controls the updating speed. A smaller $m$ increases the update sizes, speeding up the learning process. However, when $m$ is too small,  such as 2, the maximum update value can be at most one-fourth of the weight value, leading the instability and degragation in model performance.

\section{Conclusion and Discussion}\label{sec:conclusion}

In this paper, we proposed Tin-Tin, an innovative integer-based on-device training framework designed for low-power MCUs. Our approach addresses the significant challenges posed by the limited memory and computational capabilities of these devices, particularly under the lack of dedicated FPUs. By leveraging novel integer rescaling techniques, Tin-Tin efficiently manages dynamic ranges and facilitates precise weight updates using reduced-precision data types. Through comprehensive real-world experiments, we validated the effectiveness of Tin-Tin across various sensor-based learning applications. Our case studies on motor bearing fault detection and spectrum sensing demonstrated the practical benefits of Tin-Tin in real-world scenarios. Specifically, the experiments showcased Tin-Tin’s ability to significantly reduce training time, memory usage, and energy consumption compared to traditional full-precision floating-point implementations.

These results underscore the potential of Tin-Tin to support energy-efficient and sustainable ML applications on edge platforms. Our framework not only extends the battery life of sensor devices but also enhances their capability to adapt to changing environments through continuous learning. Future work will include further optimizing the framework and exploring its applicability to a broader range of edge devices and learning tasks.
\bibliographystyle{ACM-Reference-Format}
\bibliography{reference}
\end{document}